\documentclass{article} 
\usepackage{iclr2023_conference,times}

\usepackage{amsmath,amsfonts,bm}

\def\eqref#1{equation~\ref{#1}}

\def\1{\bm{1}}

\DeclareMathAlphabet{\mathsfit}{\encodingdefault}{\sfdefault}{m}{sl}
\SetMathAlphabet{\mathsfit}{bold}{\encodingdefault}{\sfdefault}{bx}{n}

\newcommand{\x}{{\boldsymbol x}}

\newcommand{\z}{{\boldsymbol z}}

\newcommand{\Ib}{{\boldsymbol I}}

\newcommand{\Db}{{\boldsymbol D}}

\newcommand{\Nc}{{\mathcal N}}

\usepackage{amssymb,amsthm}
\usepackage{hyperref}
\usepackage{amsmath}
\usepackage{graphicx}
\usepackage{wrapfig,lipsum,booktabs}

\usepackage{epsfig}
\usepackage{tikz}
\usetikzlibrary{spy}
\usepackage{algpseudocode}
\usepackage{algorithm}
\usepackage{mathrsfs}

\usepackage{nicefrac}       
\usepackage{booktabs}       

\usepackage{thmtools,thm-restate}
\usepackage{cleveref}
\usepackage{subcaption}

\usepackage{url}
\usepackage{multirow}
\usepackage{makecell}
\usepackage{color}
\usepackage{colortbl}

\usepackage{hyperref}
\usepackage{url}
\usepackage{multirow}
\usepackage{makecell}

\iclrfinalcopy
\def\eqref#1{(\ref{#1})}

\title{Pyramidal Denoising Diffusion Probabilistic Models}

\author{Dohoon Ryu\\
Department of Bio and Brain Engineering\\
Korea Advanced Institute of Science and Technology(KAIST)\\
\texttt{dh.ryu@kaist.ac.kr} \\
\AND
Jong Chul Ye\\
Kim Jaechul Graduate School of AI\\
Korea Advanced Institute of Science and Technology(KAIST)\\
\texttt{jong.ye@kaist.ac.kr} \\
}

\begin{document}

\maketitle

\begin{abstract}
Recently, diffusion model have demonstrated impressive image generation performances, and have been extensively studied in various computer vision tasks. Unfortunately, training and evaluating diffusion models consume a lot of time and computational resources. To address this problem, here we present a novel pyramidal diffusion model that can generate high resolution images starting from much coarser resolution images using a {\em single} score function trained with a positional embedding. This enables a neural network to be much lighter and also enables time-efficient image generation without compromising its performances. Furthermore, we show that the proposed approach can be also efficiently used for multi-scale super-resolution problem using a single score function.
\end{abstract}

\section{Introduction}

Diffusion models produce high quality images via reverse  diffusion processes and have achieved impressive performances in many computer vision tasks. Score-based generative models~\citep{song2020score} produce images by solving a stochastic differential equation using a score function estimated by a neural network. Denoising diffusion probabilistic models (DDPMs)~\citep{ho2020denoising,sohl2015deep}  can be considered as discrete form of score-based generative models.  
Thanks to the state-of-art image generation performance, these diffusion models have been widely investigated for various applications.

For example, 
 \citet{rombach2021high} trained a diffusion model on the latent space of a convolutional neural network (CNN)-based generative model, which enabled various of tasks. DiffusionCLIP~\citep{kim2021diffusionclip} leveraged contrastive language-image pretraining (CLIP) loss~\citep{radford2021learning}  and the denoising diffusion implicit model (DDIM)~\citep{song2020denoising} for text-driven style transfer. ILVR \citep{choi2021ilvr} proposed conditional diffusion models using  unconditionally trained score functions,  and  CCDF~\citep{chung2021come} developed
its generalized frameworks and their acceleration techniques. Also, recently proposed models~\citep{nichol2021glide,ramesh2022hierarchical} have achieved incredible performances on text-conditioned image generation and editing.

In spite of the amazing performance and flexible extensions, slow generation speed remains as a critical drawback. To resolve the problem, various approaches have been investigated. \citet{rombach2021high,vahdat2021score} trained a diffusion model in a low-dimensional representational space provided by pre-trained autoencoders. DDIM~\citep{song2020denoising} proposed deterministic forward and reverse sampling schemes to accelerate the  generation speed.
\citet{song2020improved} proposed a parameterization of covariance term to achieve better performance and faster sampling speed. \citet{jolicoeur2021gotta} used adaptive step size without any tuning. PNDM~\citep{liu2022pseudo} devised a pseudo numerical method by slightly changing classical numerical methods~\citep{sauer2011numerical} for speed enhancement. \citet{salimans2022progressive} reduced the sampling time by progressively halving the diffusion step without losing the sample quality. Denoising diffusion GANs~\citep{xiao2021tackling} enabled large denoising steps through parameterizing the diffusion process by multimodal conditional GANs.
For conditional diffusion,  a short forward diffusion steps of corrupted input can reduce the number of reverse diffusion step in SDEdit~\citep{meng2021sdedit} and RePaint~\citep{lugmayr2022repaint},
whose theoretical justification was discovered in  in CCDF \citep{chung2021come} using the stochastic contraction theory.


Alternatively, this paper addresses the slow sampling time issue in a similar manner to the method in \citet{saharia2021image} and \citet{ho2022cascaded} that refine low resolution images to high resolution using cascaded applications of multiple diffusion models. However,  in contrast to \citep{saharia2021image,ho2022cascaded}, our model does not  need to train multiple models, and can be implemented on a much lighter {\em single} architecture which results in  speed enhancement in both training and inference without compromising the generation quality.

\begin{figure*}[t]
    \centering
    \includegraphics[width=1.0\textwidth]{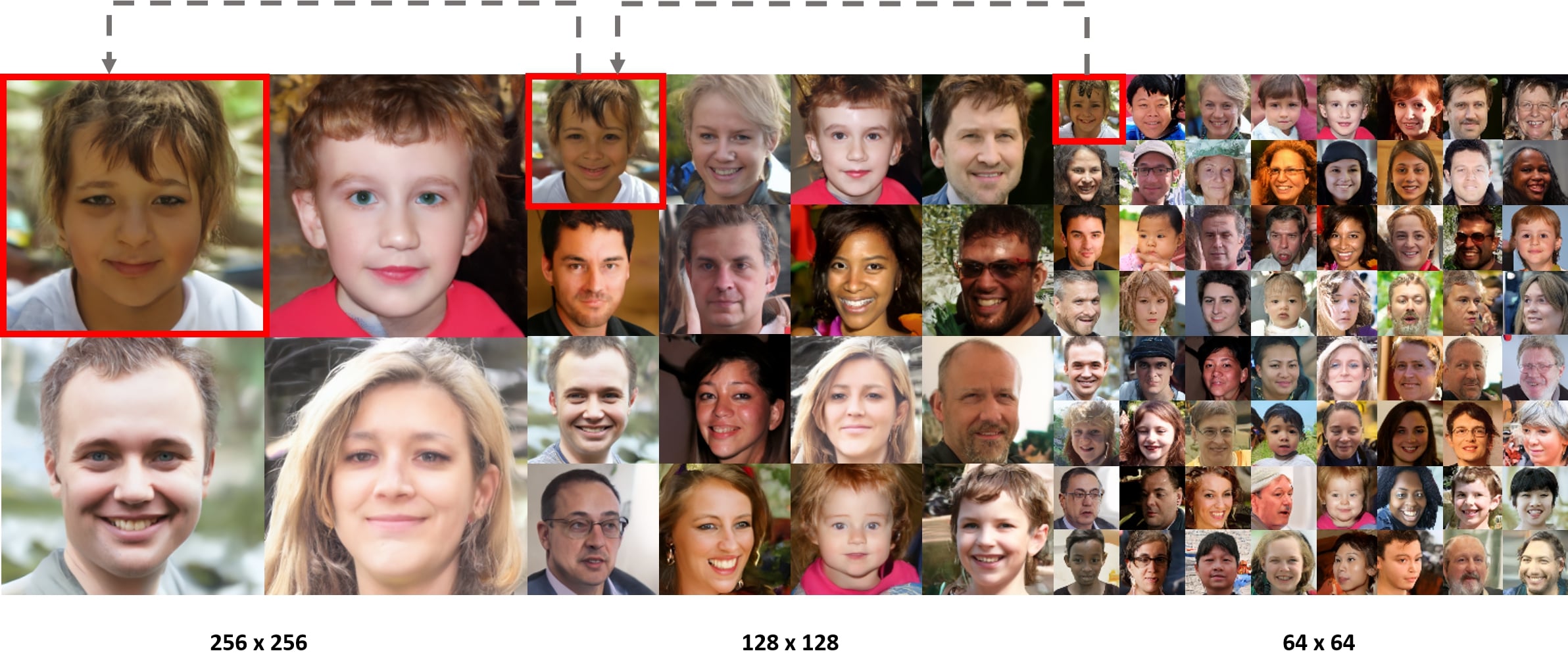}
    \caption{Progressive image generation from noises using the proposed method trained on FFHQ~\citep{choi2020stargan} dataset. Three different resolution images are generated from noise through reverse diffusion processes using a single model. In red boxes, the preservation of the semantic information at different resolution images is observed.}
    \label{fig:genimages}
\end{figure*}

Specifically,  in contrast to  the existing diffusion models that adopt encoder-decoder architecture for the same dimensional input and output, 
here we propose a new conditional training method for the score function using positional information, which gives flexibility in the sampling process of reverse diffusion. 
Specifically, our pyramidal DDPM can generate a multiple resolution images using a single score function by utilizing positional information as a condition for training and inference. 
Fig.~\ref{fig:genimages} shows the result of generated images in three different resolutions using only one model in the reverse diffusion process, which clearly demonstrates the flexibility of our method.
In addition, as a byproduct, we also demonstrate multi-scale super-resolution using a single diffusion model.

{
The contribution of this work can be summarized as following:
\begin{itemize}
    \item We propose a novel method of conditionally training diffusion model for multi-scale image generation by exploiting the positional embedding.
In contrast to the existing diffusion model, in which the latent dimension and the output dimension are the same, in our method the output dimension can be arbitrarily large compared to the latent input dimension.
    \item Using a single score network, we mitigate high computation problem and slow speed issue of reverse diffusion process using a coarse-to-fine refinement
      while preserving the generation quality.
    The key element for this is again the positional encoding as a condition for the diffusion model.
    \item We present multi-scale super-resolution which recursively refines the image resolution using a single score model.
\end{itemize}
}

\section{Background}
\label{sec:background}

\begin{figure}[t]
    \centering
    \includegraphics[width=1.0\textwidth]{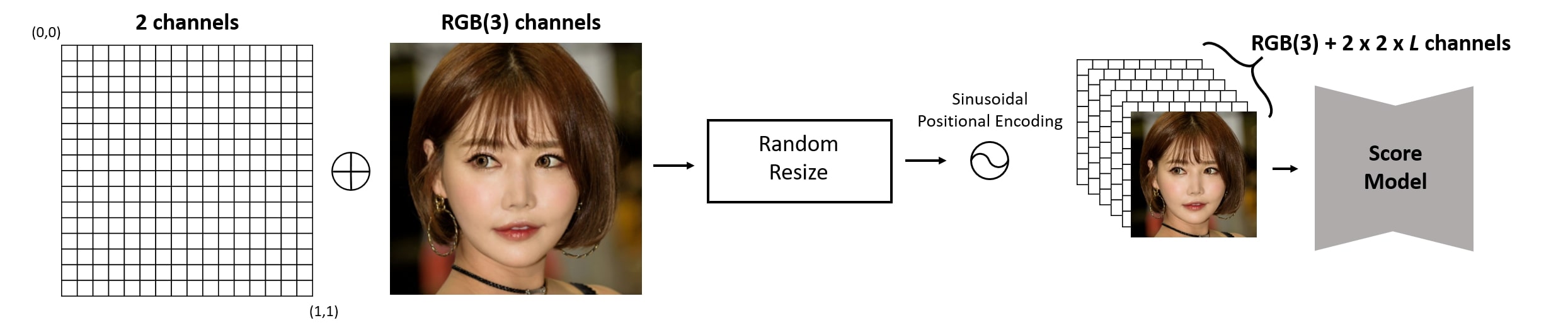}
    \caption{Our training scheme. Two dimensional coordinate information is concatenated with the input image and randomly resized to one of the target resolution. Then, two channels of coordinate values are encoded with the sine and cosine functions, and expanded to $2\times{2}\times{L}$ channels where $L$ is the degree of positional encoding.}
    \label{fig:training}
\end{figure}

\subsection{Denoising diffusion probabilistic models}
In DDPMs~\citep{ho2020denoising,sohl2015deep}, for a given data distribution $\x_0 \sim q(\x_0)$, we define a forward diffusion process $q(\x_t|\x_{t-1})$ as a Markov chain by gradually adding Gaussian noise at every time steps $t$, where $\{\beta\}_{t=0}^{T}$ is a variance schedule: 
\begin{align}
    q(\x_{T}|\x_0):=\prod_{t=1}^{T}q(\x_t|\x_{t-1}),\quad \mbox{where}\quad
    q(\x_t|\x_{t-1}):=\Nc(\x_t;\sqrt{1-\beta_t}\x_{t-1},\beta_t\Ib).
\end{align}
With well scheduled $\{\beta\}_{t=0}^{T}$, the forward process converts a data distribution to an isotropic Gaussian distribution as $t \rightarrow T$. Using the notation $\alpha_t=1-\beta_t$ and $\bar{\alpha_t}:=\prod_{s=1}^t\alpha_s$, we can sample from $q(\x_t|\x_0)$ in a closed form:
\begin{align}
    \label{eq:ddpm}
    \x_t=\sqrt{\bar{\alpha}_t}\x_0+\sqrt{1-\bar{\alpha}_t}\z,\quad \mbox{where}\quad \z\sim\Nc(\textbf{0},\Ib).
\end{align}
As the reverse of the forward step $q(\x_{t-1}|\x_t)$ is intractable, DDPM learns parameterized Gaussian transitions $p_\theta(\x_{t-1}|\x_t)$. The reverse process is defined as Markov chain with learned mean and fixed variance, starting from $p(\x_T)=\Nc(\x_T;\textbf{0},\boldsymbol{I})$:
\begin{align}
    p_\theta(\x_{0:T}):=p_\theta(\x_T)\prod_{t=1}^{T}p_\theta(\x_{t-1}|\x_{t}),\quad \mbox{where}\quad
    p_\theta(\x_{t-1}|\x_{t}):=\Nc(\x_{t-1};\bm\mu_\theta(\x_t,t),\sigma_t^2\Ib).
\end{align}
where
\begin{align}
    \bm\mu_\theta(\x_t,t):=\frac{1}{\sqrt{\alpha}_t}\Big{(}\x_t+(1-\alpha_t)\boldsymbol{s}_\theta(\x_t,t)\Big{)},\quad     \boldsymbol{s}_\theta(\x_t,t)=-\frac{1}{\sqrt{1-\bar{\alpha}_t}}\z_\theta(\x_t,t)
\end{align}
Here, $\boldsymbol{s}_\theta(\x_t,t)$ is a score function and $\z_\theta(\x_t,t)$ is trained 
 by optimizing the objective
\begin{align}
    \label{eq:objective}
    \min_{\theta} L(\theta), \quad\mbox{where}\quad L(\theta):=\mathbb{E}_{t,\x_0,\z}\Big{[}\|\z-\z_\theta(\sqrt{\bar{\alpha}_t}\x_0+\sqrt{1-\bar{\alpha}_t}\z,t)\|^2\Big{]}.
\end{align}
After the optimization, by plugging the learned score function into the generative (or reverse) diffusion process, one can simply sample  from $p_\theta(\x_{t-1}|\x_t)$ by
\begin{align}\label{eq:reverse}
    \x_{t-1}=\frac{1}{\sqrt{\alpha_t}}\Big{(}\x_t+(1-\alpha_t)\boldsymbol{s}_\theta(\x_t,t)\Big{)}+\sigma_t\z \ .
\end{align}

\subsection{Positional embedding}
Positional embedding or encoding is widely used in many recent studies in order to give locational information to deep neural networks. 
For example, \citet{devlin2018bert,dosovitskiy2020image} add position embedding to every entry of inputs in the form of trainable parameters. Unlike trainable embedding, ~\citet{tancik2020fourier} proposed a method applying sinusoidal positional encoding of the coordinate values for various tasks such as image regression, 3D shape regression, or MRI reconstruction.  As the wave is continuous and periodic, low dimensional information can be expanded to a high dimensional space of different frequencies. 
In particular, the distance between periodically encoded vectors can be easily calculated by a simple dot product, so that the relative positional information of the data is provided without any additional effort. 

The positional information is also useful for training neural network for image generation.  Authors in~\citep{anokhin2021image} modified StyleGAN2~\citep{karras2020analyzing} into a pixel-wise image generation model, which outputs RGB values from the input of 2D coordinate of pixels on image. The models of~\citep{yu2022generating,skorokhodov2021stylegan} generate continuous video frame by giving position encoded temporal information to the image generative models. The decoder in~\citep{lin2019coco} generates full images despite the model being only trained with image patches and their center coordinate. Also, ~\citet{lin2021infinitygan,ntavelis2022arbitrary} train their models through patches and their coordinates, which allows the generative model to bring out arbitrary size of images.


\begin{algorithm}[t]
\caption{Pyramidal reverse diffusion for  image generation. It starts from initialization with Gaussian noise at the coarsest resolution,
 and ends at HR. }
\label{alg:flexsampl}
\begin{algorithmic}[1]
\Require $T_f,T_s,\triangle{t_s}, \{\alpha_t^f\}_{t=1}^{T_f}, \{\alpha_t^s\}_{t=1}^{T_s}, \{\sigma_t^f\}_{t=1}^{T_f}, \{\sigma_t^s\}_{t=1}^{T_s}$

\State $\x_{T_f}^{LR} \sim \Nc(\textbf{0}, \sigma_{T_f}^f\Ib)$ \Comment{Gaussian sampling at the low resolution}
\For{$t = T_f$ to $1$} \Comment{Full reverse diffusion} \do \\
\State $\x_{t-1}^{LR} \gets \frac{1}{\sqrt{\alpha_t^f}}(\x_t^{LR} + (1-\alpha_t^f)\boldsymbol{s}(\x_t^{LR},t,pos(\boldsymbol{i}),pos(\boldsymbol{j})) + \sigma_t^f\z$
\EndFor
\State $\x_0 \gets \x_0^{LR}$
\While {$\x_0$ is $HR$} \do \\
\State $\x_0, \boldsymbol{i}, \boldsymbol{j} \gets U^{\times2}(\x_0),U^{\times2}(\boldsymbol{i}),U^{\times2}(\boldsymbol{j})$ \Comment{Upsample image and coordinate value}
\State $\x_{T_s\triangle{t_s}} \gets \sqrt{\bar\alpha_{T_s\triangle{t_s}}^s}\x_0 + \sqrt{1-\bar\alpha_{T_s\triangle{t_s}}^s}\z$ \Comment{Sample at $T_s\triangle{t_s}$}
\For{$t = T_{s}\triangle{t_s}$ to $1$} \Comment{Scaled reverse diffusion} \do \\
\State $\x_{t-1} \gets \frac{1}{\sqrt{\alpha_t^s}}(\x_t + (1-\alpha_t^s)\boldsymbol{s}(\x_t,t,pos(\boldsymbol{i}),pos(\boldsymbol{j})) + \sigma_t^s\z$
\EndFor
\EndWhile
\State \textbf{return} ${\x_0}$
\end{algorithmic}
\end{algorithm}

\section{Pyramidal denoising diffusion probabilistic models}
\label{headings}

In this section, we provide a detailed explanation of our method, called the Pyramidal Denoising Diffusion Probabilistic Models (PDDPM).
\subsection{Multi-scale score function training}

Training with multi-scale images is feasible in CNN-based models as they rely on the convolution calculation with spatially invariant filter kernels~\citep{albawi2017understanding}. Leveraging this simple but strong characteristic of the architecture, our goal is to train diffusion model such that it can understand different scale of the input by giving coordinate information as a condition. 
Specifically, we concatenate an input image and coordinate values of each pixels $(i,j)$, while $i,j \in [0,1]$ are normalized value of its location. Then, random resizing to the target resolution, 64/ 128/ 256 in our case, is applied on the merged input. The resized coordinate values are encoded with sinusoidal wave, expanded to high dimensional space, and act as conditions when training as shown in Fig.~\ref{fig:training}. 
Specifically, the positional encoding function is given by
\begin{align}
\label{eq:encoding}
    pos(\boldsymbol\gamma) &=\Big{[}\sin{(2^0\boldsymbol\gamma)},\cos{(2^0\boldsymbol\gamma)},\sin{(2^1\boldsymbol\gamma)},\cos{(2^1\boldsymbol\gamma)}\cdot\cdot\cdot\sin{(2^{L-1}\boldsymbol\gamma)},\cos{(2^{L-1}\boldsymbol\gamma)}\Big{]} \\
    \boldsymbol\gamma&=
    \begin{bmatrix}
        \gamma_1 & \gamma_2 & \cdots & \gamma_n
    \end{bmatrix}^\intercal \notag
\end{align}
where $\gamma_i \in[0,1]$ and 
$L$ denotes the degree of positional encoding and $n$ is dimension of vector. By denoting $\boldsymbol{i}=\begin{bmatrix}i_1 & i_2 & \cdots & i_N\end{bmatrix}^\intercal,\boldsymbol{j}=\begin{bmatrix}j_1 & j_2 & \cdots & j_N\end{bmatrix}^\intercal$ and $N$ referring to the dimension of $\x_t$, the training cost function in \eqref{eq:objective} can be converted as
\begin{align}
\label{eq:newobjective}
    \textit{L}(\theta) := \mathbb{E}_{t,\x_{0},\z}\Big{[}\|\z-\z_\theta(\x_t,t,pos(\boldsymbol{i}),pos(\boldsymbol{j}))\|^{2}\Big{]}.
\end{align}

Benefited from the UNet-like model structure~\citep{ronneberger2015u},  the cost function Eq.~\eqref{eq:newobjective} is invariant to all different resolutions so that the optimization can be performed with only a single network.
This simple idea of scale-free training of the score network  significantly improves the network's flexibility of sampling process which will be discussed later. Importantly, this can also alleviate the problem of slow training and low batch size problems especially when training with limited resources, the latter of which is significant for higher performance generative tasks.

\subsection{Fast reverse diffusion through pyramidal reverse sampling}

Thanks to the multi-scale score function, the sampling speed, which is the most critical disadvantage of the diffusion models, can be also made much faster compared to a single full DDPM~\citep{song2020denoising} reverse process.  
Although one may think that our method is similar to
 \citep{ho2022cascaded,jing2022subspace} which trains additional score networks for sampling in lower dimensions,
 there are several important improvement in our method.
First, our method only uses a {\em single} score model, which is trained with the positional encoding as
explained in \eqref{eq:objective}.
Second, inspired by the CCDF acceleration scheme \citep{chung2021come},
the reverse sampling process can be further accelerated 
 by using the lower-resolution reconstruction as an initialization for the next higher resolution reconstruction.

Specifically,  as shown in Algorithm ~\ref{alg:flexsampl} and Fig.~\ref{fig:flexsample}(a),
we first set two different number of time steps, $T_{f}$ and $T_{s}$. Here, 
 $T_{f}$ is a total time steps for full reverse diffusion process without any short-cut path which is applied on generating images of the lowest resolution;
$T_{s}$ is a scaled time step  for higher resolution reverse processes for fast sampling.
 More specifically, starting from low resolution ($LR$) random Gaussian noise $\x_{t=T_{f}}^{LR} \sim \Nc(\textbf{0}, \Ib)$, the full reverse diffusion process is first performed, which is much faster compared to the reverse diffusion process at the maximum resolution. Then the generated $LR$ image and its position values are upscaled twice to produce the initialization for the next resolution. 
 Then, the noises are added to the scaled image through forward diffusion with $T_s$,
 after which the reverse diffusion process is performed
 from $t=T_{s}\triangle{t_s}$ to $t=0$ until the  next higher resolution image is generated with $\triangle{t_s}\in(0,1)$. This procedure
 is recursively applied to the next higher resolution images. 
  Using this pyramidal image generation, the total
 sampling time can be significantly reduced compared to the single-resolution image generation at the highest resolution.
 Furthermore, according to the CCDF theory  \citep{chung2021come},
 the reverse diffusion process is a stochastic contraction mapping so that it reduces the estimation error from the forward diffused initialization at  exponentially fast speed.
Therefore, the required number of the reverse diffusion can be significantly reduced when a better initialization is used.
 This is why we use the previous resolution reconstruction as an initialization for the next finer resolution for further acceleration.
For theoretical details, see  \citep{chung2021come}.

\subsection{Pyramidal super resolution with stable gradient guidance}

\begin{figure}
    \centering
    \includegraphics[width=1\textwidth]{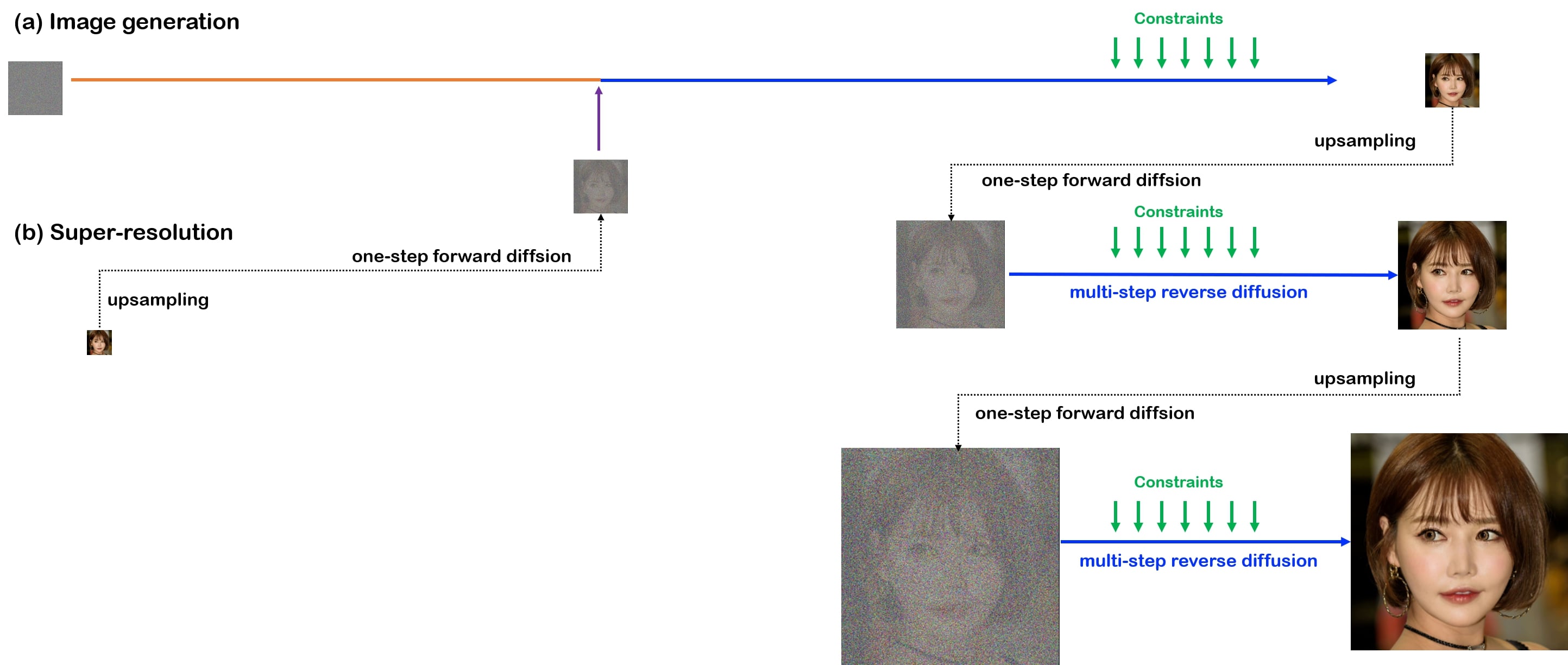}
    \caption{Proposed inference procedure for (a) image generation and (b) super-resolution.
    At the lowest resolution, full reverse diffusion is performed, which is then upscaled and forward diffused with additional noise.
   The CCDF~\citep{chung2021come} acceleration scheme is used as an acceleration scheme. For super-resolution, we imposes constraints in \eqref{eq:guide} at every step of the reverse process.}
    \label{fig:flexsample}
\end{figure}

Recall that SR3 ~\citep{saharia2021image} iteratively refines images from low to high resolution using two different score network modules. Likewise, ~\citet{ho2022cascaded,ramesh2022hierarchical,song2020improved} generate low resolution images, and uses one or two different pretrained checkpoints for super-resolution. These diffusion models showed impressively high performance, but still requires separately trained networks. However, in our case, only a single model with small number of forward and reverse processes is sufficient  as explained in the following and Fig.~\ref{fig:flexsample}(b).

\begin{algorithm}[t!]
\caption{Pyramidal reverse diffusion for super-resolution.
It starts from initialization with input LR image $\bar{\x_0}$ and ends at HR. }
\label{alg:SR}
\begin{algorithmic}[1]
\Require $\bar{\x_0},T_s,\triangle{t_s}, \{\alpha_t^s\}_{t=1}^{T_s}, \{\sigma_t^s\}_{t=1}^{T_s}$
\State $\x_0 \gets \bar{\x_0}$ \Comment{Initialization using LR input image}
\While {$\x_0$ is $HR$} \do \\
\State $\x_0, \boldsymbol{i}, \boldsymbol{j} \gets U^{\times2}(\x_0),U^{\times2}(\boldsymbol{i}),U^{\times2}(\boldsymbol{j})$ \Comment{Upsample image and coordinate value}
\State $\x_{T_s\triangle{t_s}} \gets \sqrt{\bar\alpha_{T_s\triangle{t_s}}^s}\x_0 + \sqrt{1-\bar\alpha_{T_s\triangle{t_s}}^s}\z$ \Comment{Sample at $T_s\triangle{t_s}$}
\For{$t = T_{s}\triangle{t_s}$ to $1$} \Comment{Scaled reverse diffusion} \do \\
\State $\x_{t-1} \gets \frac{1}{\sqrt{\alpha_t^s}}(\x_t + (1-\alpha_t^s)\boldsymbol{s}(\x_t,t,pos(\boldsymbol{i}),pos(\boldsymbol{j})) + \sigma_t^s\z$
\State $\hat \x_0(\x_t):=\frac{1}{\sqrt{\bar\alpha_t^s}}\left(\x_t+(1-\bar\alpha_t^s)\boldsymbol{s}_\theta(\x_t,t,pos(\boldsymbol{i}),pos(\boldsymbol{j})\right)$
\State $\x_{t-1} \leftarrow \x_{t-1}- \lambda\nabla_{\x_t}\left\|\Db \hat \x_0(\x_t) -\bar{\x}_0\right\|_2^2$
\EndFor
\EndWhile
\State \textbf{return} ${\x_0}$
\end{algorithmic}
\end{algorithm}

Here, the reverse step can be guided towards the target by subtracting the gradient to the intended direction as suggested in ~\citep{ho2022video,dhariwal2021diffusion,avrahami2022blended}. Specifically, starting from \eqref{eq:ddpm}, the denoised prediction of $\x_0$ given $\x_t$ is first computed by
\begin{align}
\hat \x_0(\x_t):=\frac{1}{\sqrt{\bar\alpha_t}}\left(\x_t+(1-\bar\alpha_t)\boldsymbol{s}_\theta(\x_t,t,pos(\boldsymbol{i}),pos(\boldsymbol{j})\right)
\end{align}
which is mixed with the reverse diffusion samples $\x_{t-1}$ in \eqref{eq:reverse} in terms of additional gradient to update the sample:
\begin{align}
\label{eq:guide}
\x_{t-1} \leftarrow \x_{t-1}- \lambda\nabla_{\x_t}\left\|\Db \hat \x_0(\x_t) -\bar{\x}_0\right\|_2^2
\end{align}
where $\Db$ denotes the down-sampling operator and $\bar\x_0$ is the low-resolution measurement.
Additionally, we use the CCDF acceleration scheme \citep{chung2021come} similar to the aforementioned
pyramidal image generation. 
One difference from the image generation
is that even at the coarsest resolution, the full reverse sampling at the coarsest level is not necessary as the lower-resolution
measurement can be utilized as an initialization after upsampling. See  Algorithm ~\ref{alg:SR} and Fig.~\ref{fig:flexsample}(b).

\begin{figure*}[t]
    \centering
    \includegraphics[width=0.9\textwidth]{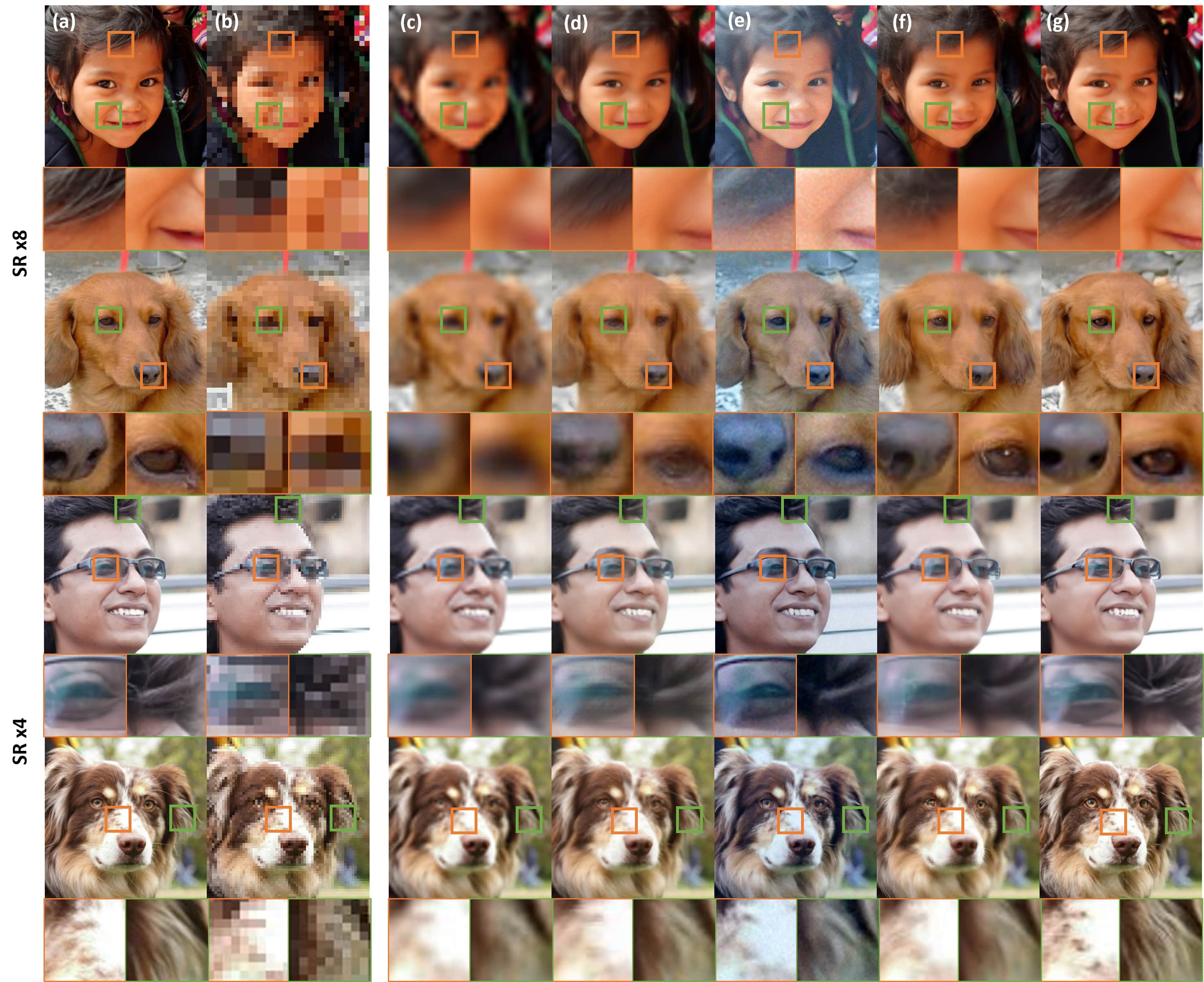}
    \caption{Result of super-resolution on FFHQ and AFHQ-dog dataset. Upper row shows the results of $\times$8 SR and bottom row is $\times$4 result. (a) Ground Truth, (b) low resolution images, the results by  (c) cubic interpolation, (d) SRGAN, (e) SR3, (f) ILVR, and (g) the proposed method.}
    \label{fig:SR_result}
\end{figure*}

\section{Experiments}
\label{sec:exp}

\subsection{Datasets and implementation}

We trained our model using FFHQ 256$\times$256~\citep{choi2020stargan}, LSUN-Church 256$\times$256~\citep{yu15lsun} datasets for 1.2M iterations and 500k for AFHQ-Dog 256$\times$256 dataset with batch size of 48/12/3 for 64/128/256 image resolution. FFHQ and LSUN-Church dataset were used to evaluate generation performances while AFHQ-Dog dataset was used for super-resolution. Our model is based on an improved version of DDPM~\footnote{\url{https://github.com/openai/improved-diffusion}}. To alleviate the memory and training speed problem in the limited resource environment using one GeForce 1080 Ti for all model training, we set the number of channels and residual blocks to  64 and 1, respectively, which is a  very  light model compared to the usual settings. Also we used 1000 diffusion steps for all training. We used Adam optimizer of 0.0001 learning rate. The degree of positional encoding $L$ in~\eqref{eq:encoding} was set to 6. For the base setting of the inference, we set $T^f=1000$ and $T^s=100$.

\begin{table}[!hbt]
\centering
\caption{
{
Frechet Inception distance (FID$\downarrow$) comparison using 10k generated images. Sampling speed was also calculated by comparing with a full reverse diffusion process. Speed for the baselines
were calculated using 100 sampling steps (gray-colored cells). The models are trained for 1.2M iterations for fair comparison.
}
}
\resizebox{0.8\textwidth}{!}{
\begin{tabular}{cccccccccc}
\toprule
{\textbf{Dataset}} & {\textbf{Method}} & {\textbf{10}} & {\textbf{20}}& {\textbf{50}}& {\textbf{100}} & {\textbf{200}} & {\textbf{-}} & {\textbf{Params}} & {\textbf{Speed}} \\
\midrule
\multirow{5}{*}{\textbf{FFHQ}} & {FON} & {26.71}& {18.91}& {16.70}& \cellcolor{gray!15}{16.13} & {16.22} & {-}& \multirow{5}{*}{114M}&\cellcolor{gray!15}{$\times{9.13}$} \\
                & {DDIM} & {37.87}& {27.71}& {20.45}& \cellcolor{gray!15}{18.30}& {16.85}& {-} &&\cellcolor{gray!15}{$\times{10.1}$}\\
                & {S-PNDM} & {28.59}& {22.62}& {18.13}& \cellcolor{gray!15}{16.71}& {16.48}& {-} &&\cellcolor{gray!15}{$\times{9.92}$}\\
                & {F-PNDM} & {25.10}& {19.16}& {16.30}& \cellcolor{gray!15}{16.09}& {16.32}& {-}&&\cellcolor{gray!15}{$\times{9.20}$}\\
                \cmidrule{2-10}
                & {Ours} & {-}& {-}& {-}& {-}& {-}& \cellcolor{gray!15}{\textbf{15.78}}& {16M}&\cellcolor{gray!15}{\textbf{$\times$18.1}} \\
\midrule
\multirow{5}{*}{\makecell{\textbf{LSUN}\\\textbf{Church}}} & {FON} & {28.93} & {27.44}& {27.11}& {25.29}& {25.72} & {-}& \multirow{5}{*}{114M} & \multirow{6}{*}{-} \\
                & {DDIM} & {26.72}& {26.65}& {25.89}& {24.36}& {25.30}& {-}\\
                & {S-PNDM} & {29.79}& {27.15}& {26.82}& {26.93}& {26.33}& {-}\\
                & {F-PNDM} & {33.80}& {27.07}& {26.95}& {25.94}& {25.78}& {-}\\
                \cmidrule{2-9}
                & {Ours} & {-}& {-}& {-}& {-} & {-}& {\textbf{14.07}}& {16M}\\
\bottomrule
\end{tabular}
}
\label{tab:comparison-FID}
\end{table}

\subsection{Image generation}

{
We compared the quality of generated images among different reverse diffusion methods devised for fast sampling. Specifically, we trained a new model for numerical method (FON), DDIM, S-PNDM and F-PNDM~\citep{liu2022pseudo}~\footnote{\url{https://github.com/luping-liu/PNDM}} for 1.2M iteration with batch size of 8. As the model consists of 7 times larger parameters than ours, the training was performed on Quadro RTX 6000 to handle the low batch size problem. This means that the baselines are trained on better condition than ours. The model uses linear noise schedule and we chose 10, 20, 50, 100 and 200 steps for the total sampling steps. When sampling from ours, we set $\triangle{t_s}$ = 0.3 so that all samplings procedure of higher resolution images can be done in only 30 steps. The sampling speed was calculated by comparing to the full diffusion step of the larger model.
}

We sampled 10k images from each method and evaluated visual quality using Frechet Inception distance based on
\verb+pytorch-fid+\footnote{\url{https://github.com/mseitzer/pytorch-fid.git}}.
The result on Table~\ref{tab:comparison-FID} shows that our method  produces superior results compared to the baselines despite the faster sampling speed with much smaller architecture. This implies that the model focuses on generating realistic images at lower resolution and adding fine details at higher resolution.
 Also the results on LSUN-Church dataset show that the training of the baselines was incomplete, whereas our method can produce qualitative results even with much  limited resources.

\begin{wraptable}[]{r}{0.45\textwidth}
\centering
\caption{
Self-Comparison results. $T_f$ is fixed to 100 which is shorter than the setting in Table~\ref{tab:comparison-FID}. FID score and sampling speed are measured in the same way as in Table~\ref{tab:comparison-FID}.}
\setlength{\tabcolsep}{0.5em}
\resizebox{0.45\textwidth}{!}{
\begin{tabular}{cccccc}
\toprule
{\textbf{$\triangle{t_s}$}} & {0.1} & {0.2} & {0.3} & {0.4} & {0.5}  \\
\midrule
{FFHQ} & 23.91 & 22.42 &22.11 & 21.91 & \textbf{21.87} \\
{LSUN-Church} & 27.39 & 18.10 & 15.38 & \textbf{14.57} & 14.63 \\
\midrule
Speed & {\textbf{$\times$82.4}} & {$\times{50.3}$} & {$\times{36.1}$} & {$\times{28.1}$} & {$\times{22.9}$} \\
\bottomrule
\end{tabular}
}
\vspace{0.2em}
\label{tab:self-comparsion}
\end{wraptable}
We can further speed up the reverse process by re-spacing total diffusion steps of lower dimension images. We changed $T^f=T^s=100$ and studied the effect of $\triangle{t_s}$ on image quality. 
By fixing total diffusion steps same for all resolution, we compared FID score on generated image when $\triangle{t_s}$ value changes from 0.1 to 0.5. 
The sampling speed computation was done similar  as done in Table~\ref{tab:comparison-FID}(a). Setting $T^f=100$ and $\triangle{t_s}=0.1$ increased the speed to be 82.4 times faster than original full diffusion process, while generating convincing images.  Table~\ref{tab:self-comparsion}(b) shows that large sampling steps on high resolution produce higher FID scores, but the sampling speed decreases proportionally to $\triangle{t_s}$. Also, compared to the baselines in Table~\ref{tab:comparison-FID}(a), our method produced comparable results to the baseline despite the faster sampling speed.

\subsection{Super resolution}

Experiment was performed on two SR factors: $\times$4, and $\times$8.
 We have tested on FFHQ, AFHQ-Dog 256$\times$256 dataset with our models and other baselines: SRGAN, SR3, ILVR and traditional bicubic upsampling method  for comparison.
 For SRGAN and SR3, there was no checkpoint for appropriate evaluation, so the models were trained on 32$\to$256, 64$\to$256 for each dataset. Additionally, the original work of SR3~\citep{saharia2021image} cascaded two $\times$4 upsamplers for super-resolution, but in this case we reported on the result of super-resolution with one model for each SR factors. We made a comparison through FID$\downarrow$, LPIPS$\downarrow$, PSNR$\uparrow$ and SSIM$\uparrow$. Here, LPIPS \footnote{\url{https://github.com/richzhang/PerceptualSimilarity}} was calculated using the open source of the perceptual similarity of VGG~\citep{simonyan2014very}.
  As shown in Table~\ref{tab:SR}, our method produced the best result on FID and LPIPS for most of the cases, but not on PSNR and SSIM. 
Although SRGAN provides the best PSNR and SSIM values, it contains  many image artifacts and the results of our method are more realistic as show in Fig.~\ref{fig:SR_result}.

\begin{table}[t]
\centering
\setlength{\tabcolsep}{0.2em}
\caption{
FID($\downarrow$), LPIPS($\downarrow$), PSNR($\uparrow$), SSIM($\uparrow$) score evaluation on super-resolution task. \textbf{Bold face}: best, \underline{underline}: second best. (*:Unofficial re-implementation.)}
\resizebox{1\textwidth}{!}{
\begin{tabular}{ccccccccccccccccc}
\toprule
{} & \multicolumn{8}{c}{\textbf{FFHQ} ($\bf 256 \times 256$)} & \multicolumn{8}{c}{\textbf{AFHQ} ($\bf 256 \times 256$)} \\
\cmidrule(lr){2-9}
\cmidrule(lr){10-17}
{\textbf{SR facotor}} & \multicolumn{4}{c}{\textbf{$\times$4}} & \multicolumn{4}{c}{\textbf{$\times$8}} &\multicolumn{4}{c}{\textbf{$\times$4}} &\multicolumn{4}{c}{\textbf{$\times$8}} \\
\cmidrule(lr){2-5}
\cmidrule(lr){6-9}
\cmidrule(lr){10-13}
\cmidrule(lr){14-17}
{\textbf{Method}} & {FID} & {LPIPS} & {PSNR} & {SSIM} & {FID} & {LPIPS} & {PSNR} & {SSIM} & {FID} & {LPIPS} & {PSNR} & {SSIM} & {FID} & {LPIPS} & {PSNR} & {SSIM}\\
\midrule
Bicubic &125.7 &0.278 & 28.84 & 0.851 &151.3 &0.446& 24.73 & 0.703 &35.34 &0.280& \underline{29.10} & \underline{0.816} &69.47 &0.429& 25.34 & \underline{0.671} \\
\addlinespace[0.2em]
SRGAN &\textbf{46.85} &\underline{0.204}& \textbf{29.45} & \underline{0.857} &71.69 &0.296& \textbf{26.20} & \textbf{0.753} &\underline{22.61} &0.247& \textbf{29.48} & \textbf{0.826} & 45.35 &0.329 & \textbf{25.91} & \textbf{0.688}\\
\addlinespace[0.2em]
{SR3*} &59.31 &0.291 & 20.56 & 0.725 &91.57 &0.404 & 18.90 & 0.612 &27.89 &0.345 & 20.15 & 0.730 &\underline{35.09} &0.401 & 18.85 & 0.622\\
\addlinespace[0.2em]
{ILVR} &54.73 &0.224 & \underline{29.15} & 0.851 &\underline{71.47} & \underline{0.295} & \underline{24.92} & 0.712 &26.61 &\textbf{0.234} & 28.76 & 0.800 &37.39 &\underline{0.326} & 25.10 & 0.650\\
\addlinespace[0.2em]
{Ours} &\underline{49.06} &\textbf{0.192} & 28.70 & \textbf{0.860} &\textbf{66.80} &\textbf{0.289} & 24.83 & \underline{0.725} &\textbf{21.62} &\underline{0.237} & 27.74 & 0.796 &\textbf{33.14} &\textbf{0.311} & \underline{25.38} & 0.649 \\
\bottomrule
\end{tabular}
}
\label{tab:SR}
\end{table}

\subsection{Ablation study}
\label{sub:ablation}

\begin{figure}[t]
\centering
    \includegraphics[width=\textwidth]{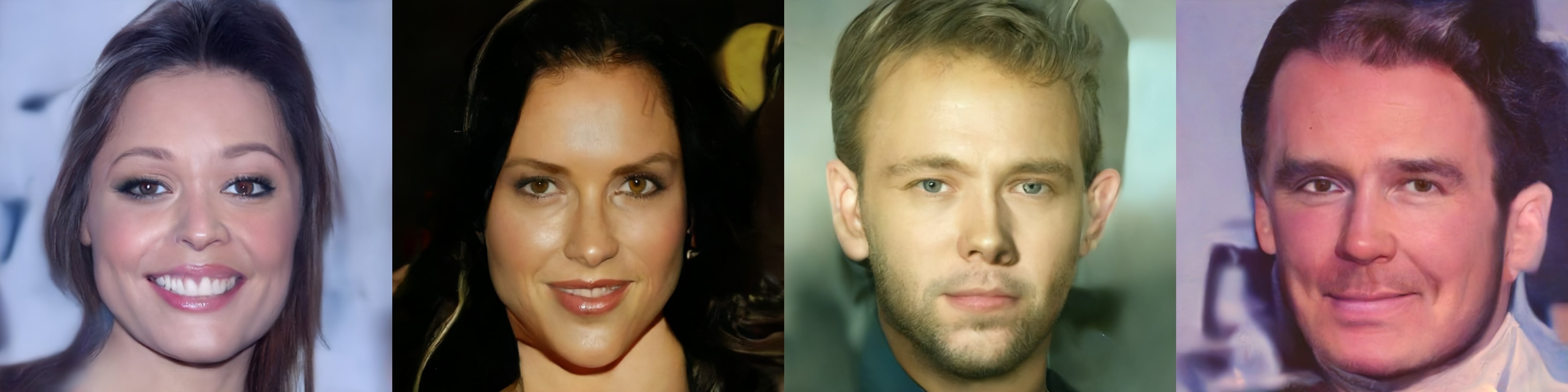}
    \caption{\label{fig:1024} Generated images of at full resolution (1024$\times$1024) by our method trained with only 256$\times$256, 512$\times$512 patches. The model had never seen full resolution image.}
\end{figure}

We performed an ablation study on pyramidal super-resolution and positional encoding. The ablation study on super-resolution was done by removing some steps of the resolution while up-sampling, in order to see the effect of pyramidal super-resolution. For positional encoding, we first trained a new model with multi-scale images without positional encoding. Second, we trained a score model with patches of $256\times256$ and $512\times512$ images for generation of $1024\times1024$ images.

\begin{wraptable}[]{r}{0.45\textwidth}
\caption{
FID($\downarrow$), LPIPS($\downarrow$) scores of ablation study on super-resolution tasks.}
\centering
\setlength{\tabcolsep}{0.5em}
\resizebox{0.4\textwidth}{!}{
\begin{tabular}{l@{\hskip 15pt}cccc}
\toprule
{} & \multicolumn{2}{c}{\textbf{FFHQ}} & \multicolumn{2}{c}{\textbf{AFHQ}} \\
\cmidrule(lr){2-3}
\cmidrule(lr){4-5}
{\textbf{Method}} & {FID} & {LPIPS} & {FID} & {LPIPS} \\
\midrule
{$32\rightarrow256$} &{72.78}&{0.308}&{45.99}&{0.340} \\
{$32\rightarrow64\rightarrow256$} &{68.85}&{0.301}&{44.25}&{0.340} \\
{$32\rightarrow128\rightarrow256$} &{68.33}&{0.291}&{36.80}&{0.328} \\
{Original} &{\textbf{66.80}}&{\textbf{0.289}}&{\textbf{33.14}}&{\textbf{0.311}} \\
\bottomrule
\end{tabular}
}
\label{tab:SR_ablation}
\end{wraptable}
Pyramidal super-resolution was evaluated by comparing the original method $32\rightarrow64\rightarrow128\rightarrow256$ with the up-sampling procedures of $32\rightarrow64\rightarrow256$, $32\rightarrow128\rightarrow256$ and $32\rightarrow256$. The evaluation was performed on both FFHQ and AFHQ-Dog dataset. The result in Table~\ref{tab:SR_ablation} shows that using all the resolution step of pyramidal super-resolution has achieved the best result. The CCDF ~\citep{chung2021come} acceleration scheme of every resolution has shown  quality improvement of reconstructed images.

\begin{wrapfigure}[]{r}{0.45\textwidth}
\centering
    \includegraphics[width=0.4\textwidth]{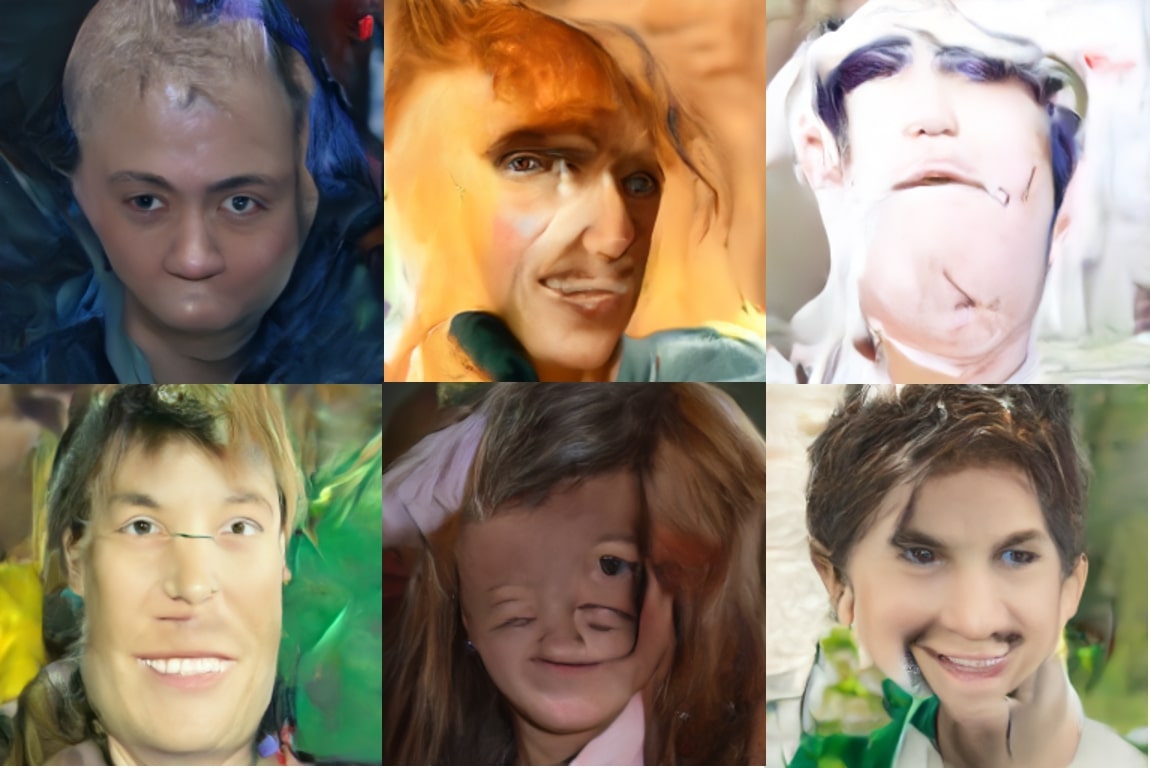}
    \caption{\label{fig:no_pos}Images generated from a pyramidal DDPM without positional encoding. Facial features are not in proper location and some are crushed.}
\end{wrapfigure}
For the first ablation experiment on the positional encoding, we trained the model with the same experiment setting except for the positional encoding. It was trained for 800k iterations. Fig.~\ref{fig:no_pos} shows the result of the first experiment. Odd looking faces with multiple eyes, crushed face or dislocated facial features were generated. The result confirmed the importance of positional encoding.

Experiment on patch-wise learning for very high-resolution image generation was performed using one NVIDIA RTX 3090. The model was trained for 400k iterations with batch size of 6/2 on 256/512 scaled image patches of CelebA-HQ~\citep{karras2017progressive} dataset. When training, random resizing and cropping were used. Sampling was performed by setting $T^f=T^s=100$. As seen in Fig.~\ref{fig:1024}, 
although the score model had never seen a maximum resolution input, $1024\times1024$ images were generated through our fast pyramidal image generation.

\section{Conclusion}

In this work, we proposed a novel Pyramidal DDPM which is trained with conditions on the positional information. We showed that this simple change improves the speed of reverse diffusion process and the performance of super-resolution and image generation. We also tested the effect of positional encoding by additional ablation experiments. Especially, without positional encoding, the model lost the ability to predict proper images at different resolution. Also patch-wise training further improved the flexibility of the score model, generating very high resolution images without using full resolution images. Given the significant advantages from a simple modification, we believe that our method may further mitigate high computation problems in the diffusion models and be used for many other applications.

\bibliography{iclr2023_conference}
\bibliographystyle{iclr2023_conference}

\newpage
\appendix

\section{Additional results}

\subsection{Additional samples}
\begin{figure}[b]
\centering
    \includegraphics[width=\textwidth]{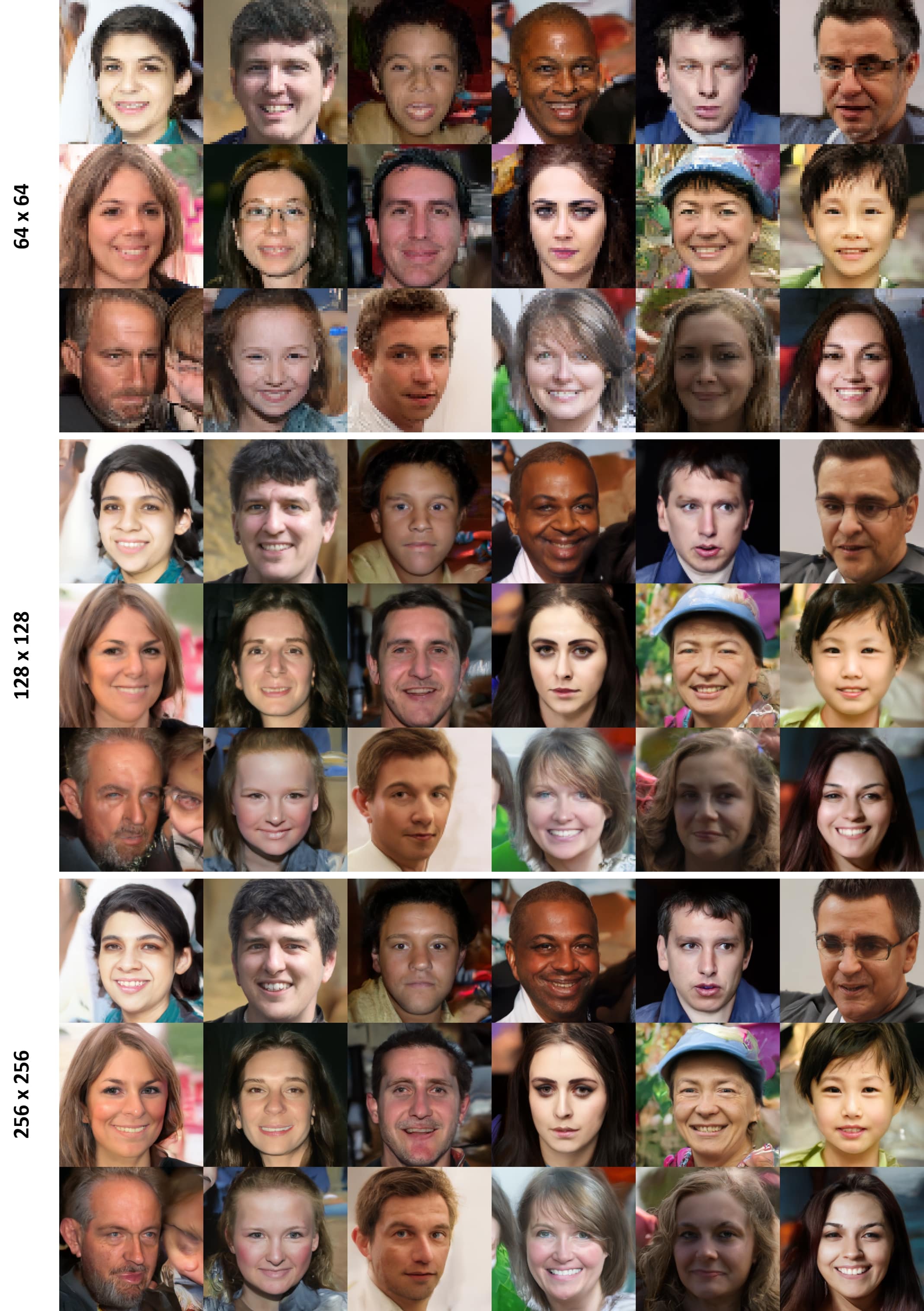}
    \caption{\label{fig:ffhq_add} Additional generated images by our method trained from FFHQ dataset. Resolution increases starting from top to bottom row.}
\end{figure}

\begin{figure}[t]
\centering
    \includegraphics[width=\textwidth]{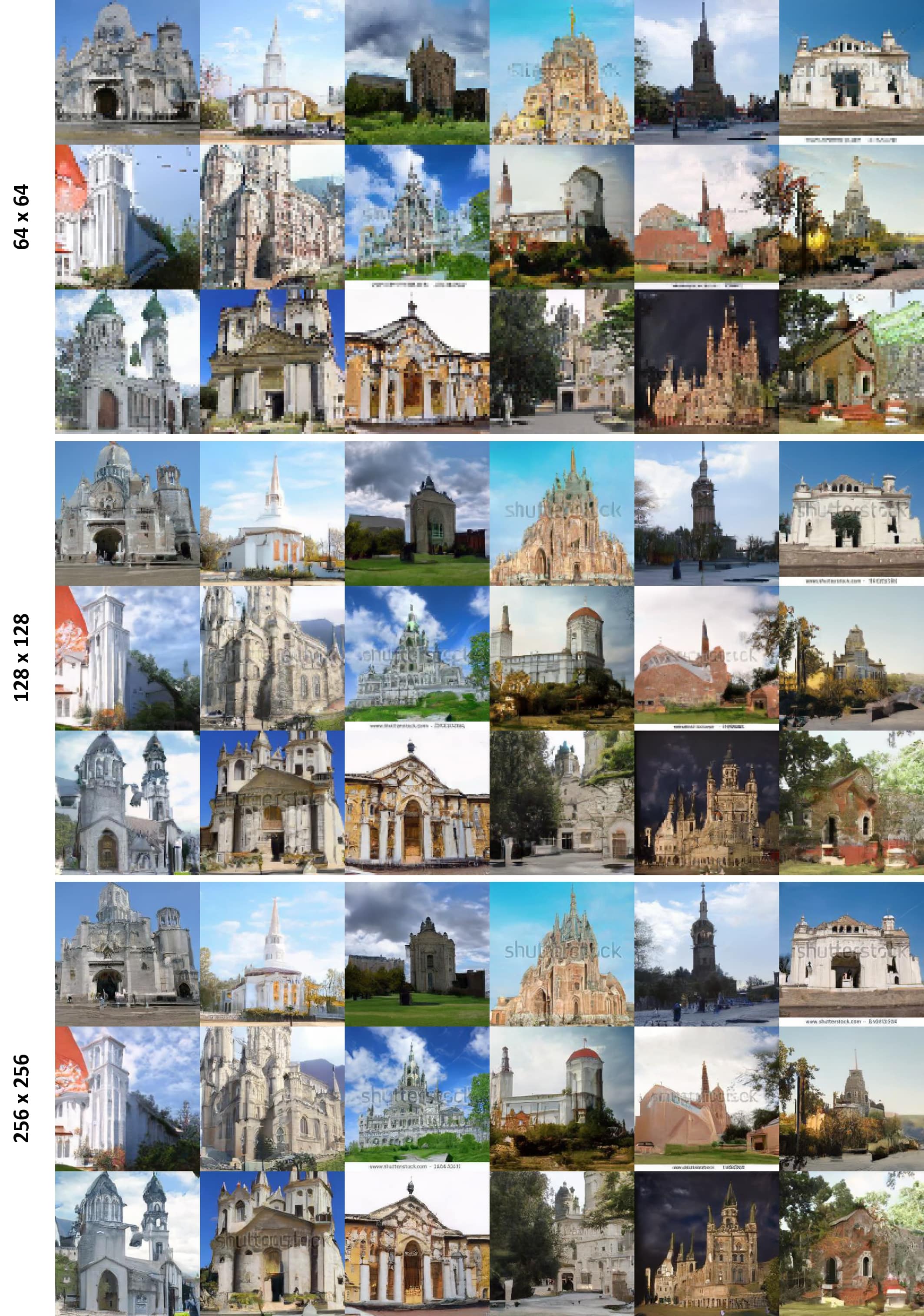}
    \caption{\label{fig:church_add} Additional  generated images by our method  trained from LSUN-Church dataset. Resolution increases starting from top to bottom row.}
\end{figure}

In this section, we provide additional samples of different resolution generated from a single score model on FFHQ and LSUN-Church dataset in Fig.~\ref{fig:ffhq_add} and Fig.~\ref{fig:church_add}.

\subsection{Super-resolution: Intermediate results}
\begin{figure}[t]
\centering
    \includegraphics[width=\textwidth]{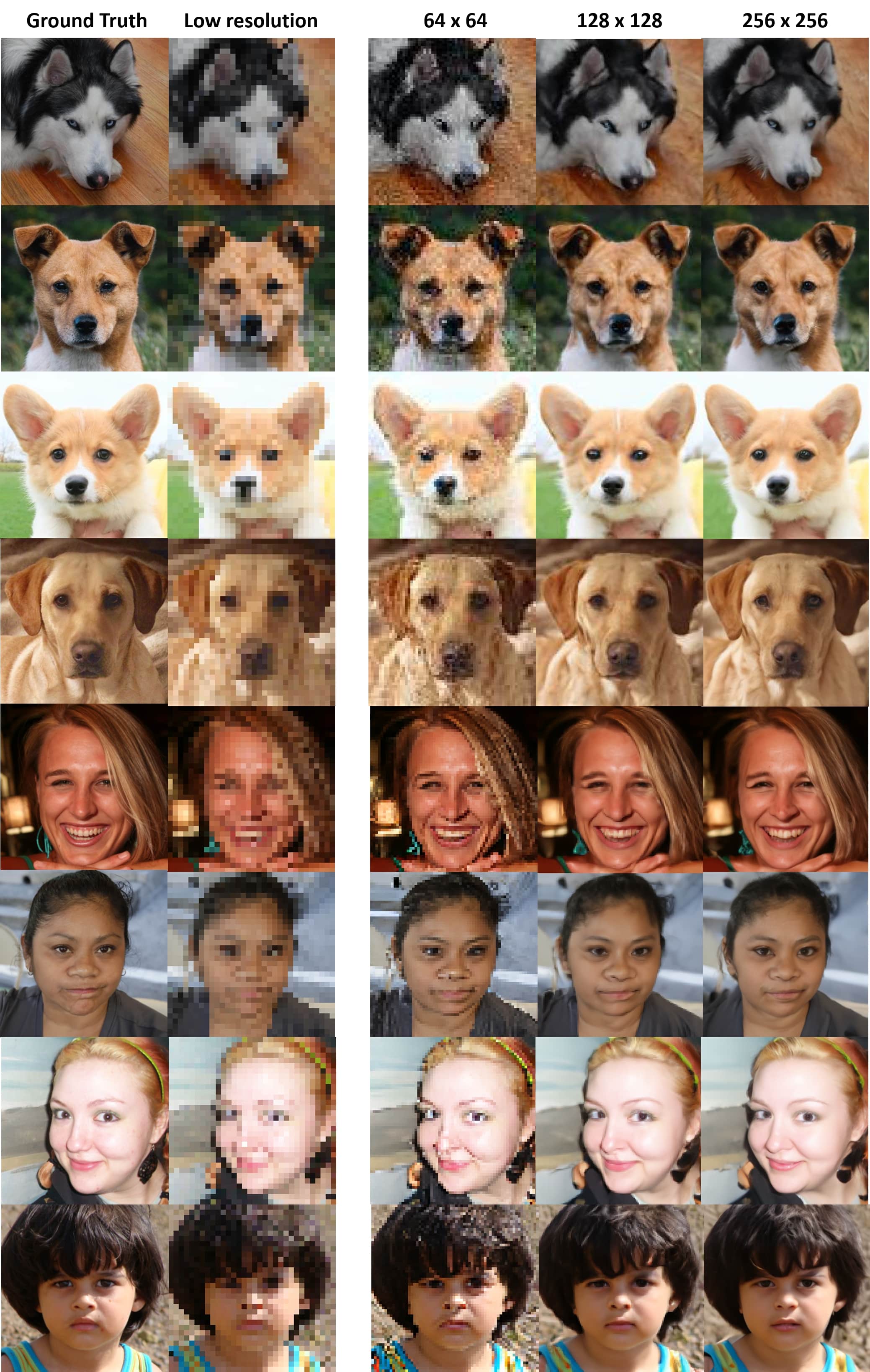}
    \caption{\label{fig:SR_gradual} Intermediate results of pyramidal super-resolution on FFHQ and AFHQ-dog dataset.}
\end{figure}

In this section, we provide intermediate results on pyramidal super-resolution task. Fig.~\ref{fig:SR_gradual} shows up-scaled images of each resolution starting from 32$\times$32 on FFHQ and AFHQ-dog dataset.

\subsection{Super-resolution: Self-comparison results}
\begin{figure}[t]
\centering
    \includegraphics[width=\textwidth]{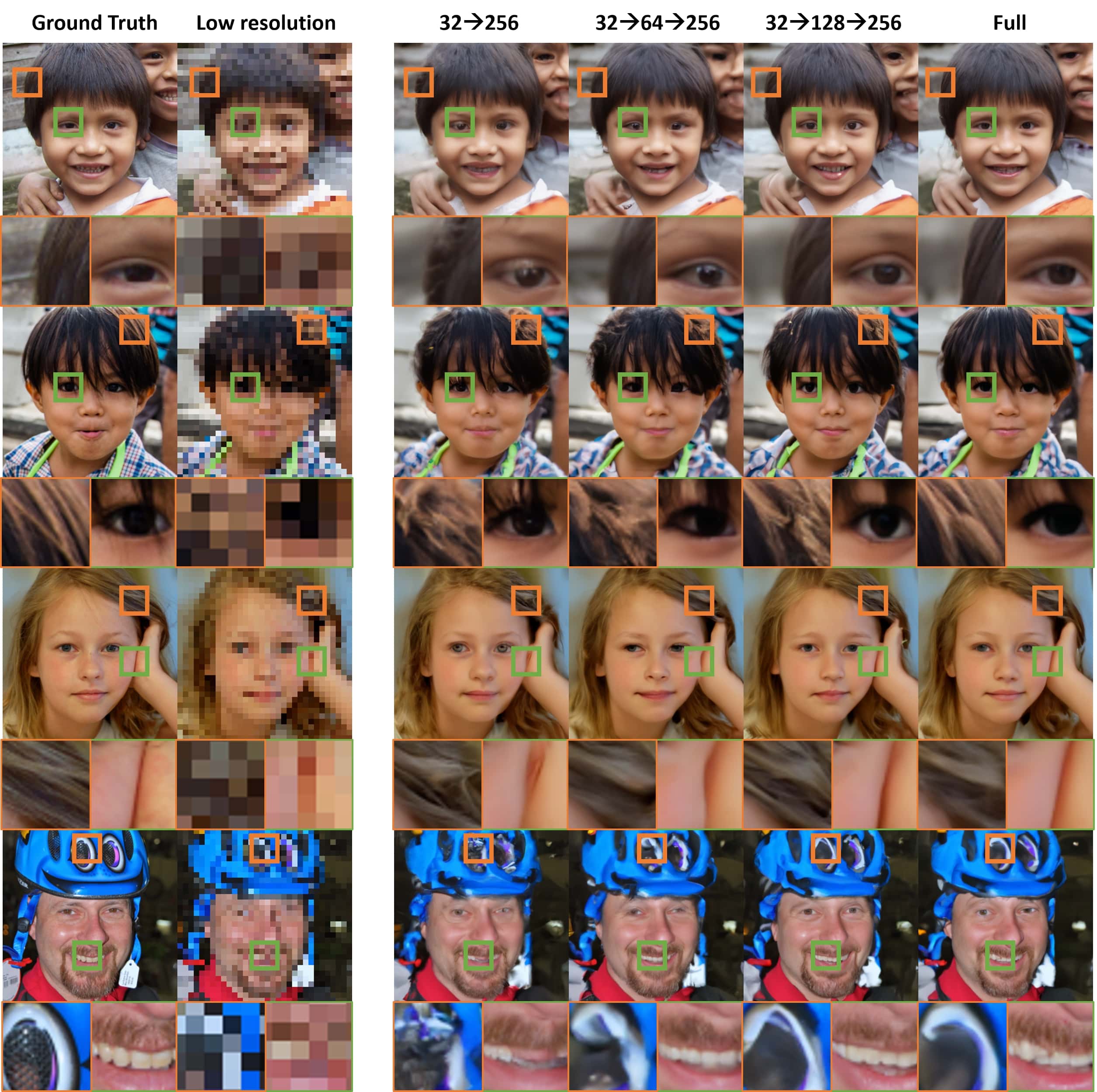}
    \caption{\label{fig:SR_ffhq_add} Visualization of the results of $\times8$ super-resolution in Table~\ref{tab:SR_ablation} on FFHQ dataset.}
\end{figure}

\begin{figure}[t]
\centering
    \includegraphics[width=\textwidth]{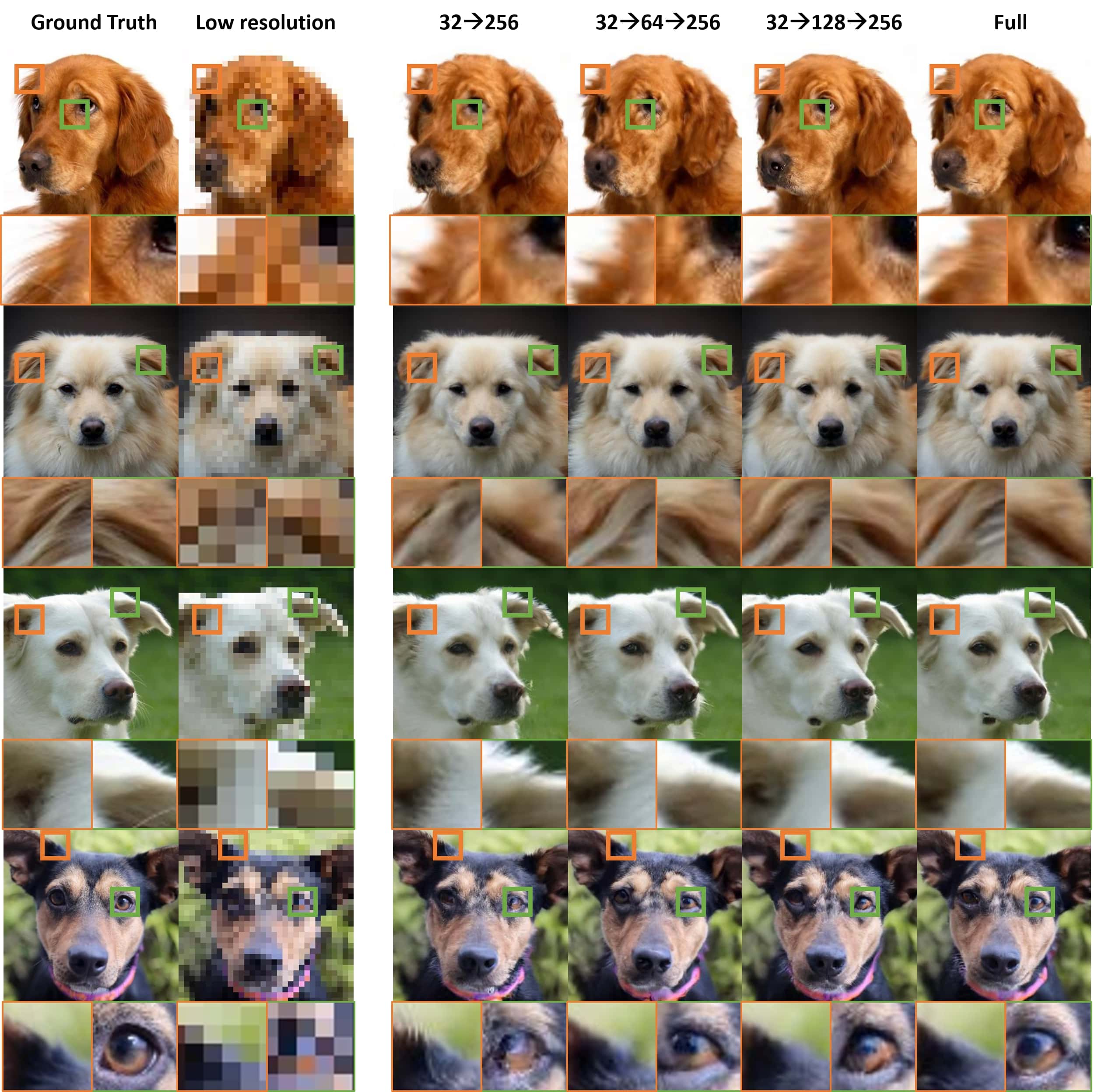}
    \caption{\label{fig:SR_afhq_add} Visualization of the results of $\times8$ super-resolution in Table~\ref{tab:SR_ablation} on AFHQ-dog dataset.}
\end{figure}

We visualized the result in Table~\ref{tab:SR_ablation} in Fig.~\ref{fig:SR_ffhq_add} and Fig.~\ref{fig:SR_afhq_add}. Both of the result show that direct path from the low resolution image to the maximum resolution produce more details in the images. This implies the effect of iterative refinement method is not trivial.

\section{Experimental details}
\subsection{Comparison methods}
\subsubsection{Image generation}

\paragraph{Pyramdial DDPM}
{
We implemented Pyramdial DDPM based on the original code of improved diffusion model~\citep{nichol2021improved}~\footnote{\url{https://github.com/openai/improved-diffusion}}. We used the basic setting of the model for generation of $64\times{}64$ images, which consists of 64 number of convolution channels and 1 block for every residual connection. The model was trained using on GeForce 1080 Ti.
}

\paragraph{FON, DDIM, PNDM}
{
We used the score function from PNDM~\citep{liu2022pseudo} official github repository~\footnote{\url{https://github.com/luping-liu/PNDM}} as a baseline. The model uses 128 number of convolution channels for every block and 2 blocks for each residual connection. Attention resolution for every layer is fixed to 16. Adam optimizer with learning rate of 0.0002 was used and the exponential moving average update is used for the inference checkpoint. This setting is exactly the same as the original setting when training LSUN-Church dataset by the author. Both the model of LSUN-Church and FFHQ dataset was trained for 1.2M iterations for fair comparison with our model. FON, DDIM, S-PNDM and F-PNDM methods are used. The model was trained using Quadro RTX 6000 to use larger batch size.
}

\subsubsection{Super resolution}
\paragraph{SRGAN}
We trained a model from scratch on FFHQ and AFHQ-dog dataset using a code from SRGAN repository~\footnote{~\url{https://github.com/leftthomas/SRGAN}}. The model was trained for 500k iteration for each dataset. The model was trained with batch size of 8, 0.001 learning rate, 0.9 $\beta_1$ and 0.999 $\beta_2$ using Adam optimizer. We used one GeForce 1080 Ti for training.
\paragraph{SR3}
We used unofficial implementation for SR3 model from github repository\footnote{~\url{https://github.com/Janspiry/Image-Super-Resolution-via-Iterative-Refinement}}. The official model trained two separate model for $\times4$ super-resolution, but we trained a single upsampler for $32\rightarrow256$ and $64\rightarrow256$. The training was done for 500k iteration with batch size of 2 for each dataset. The model consists of inner channels starting from 64 and 2 residual blocks. It was trained with 0.0001 learning rate using Adam optimizer. We used one GeForce 1080 Ti for training.
\paragraph{ILVR}
We used pretrained network for both dataset from ILVR repository~\footnote{\url{https://github.com/jychoi118/ilvr_adm}}. It was trained for 1M iterations for FFHQ, and 500k for AFHQ-dog. The model architecture and training setting were same as the one we used for DDIM and DDPM sampling.

\end{document}